\documentclass{article}

    \usepackage[final, nonatbib]{neurips_2019}

\usepackage[utf8]{inputenc} 
\usepackage[T1]{fontenc}    
\usepackage{hyperref}       
\usepackage{url}            
\usepackage{booktabs}       
\usepackage{amsfonts}       
\usepackage{nicefrac}       
\usepackage{microtype}      
\usepackage{amsmath}
\usepackage[ruled,vlined]{algorithm2e}

\usepackage{tabto}
\usepackage{amssymb}
\usepackage{mathdots}
\usepackage[pdftex]{graphicx}
\usepackage{fancyhdr}
\usepackage{multicol}
\usepackage{bm}
\usepackage{listings}
\PassOptionsToPackage{usenames,dvipsnames}{color}  
\usepackage{pdfpages}
\usepackage{algpseudocode}
\usepackage{enumerate}
\usepackage{tikz}
\usepackage{enumitem}
\usepackage[T1]{fontenc}
\usepackage{inconsolata}
\usepackage{framed}
\usepackage{tabulary}
\usepackage[thinlines]{easytable}
\usepackage{amsmath}  
\usepackage{wrapfig}  
\usepackage{hyperref} 
\usepackage{float}

\newcommand{\bfz}{\mathbf{z}}
\newcommand{\bfx}{\mathbf{x}}

\title{SGVAE: Sequential Graph Variational Autoencoder}

\author{Bowen Jing \\
  Computer Science Department \\
  Stanford University \\
  \texttt{bjing@stanford.edu}
  \And
  Ethan A. Chi \\
  Computer Science Department \\
  Stanford University \\
  \texttt{ethanchi@cs.stanford.edu}
  \AND
  Jillian Tang \\
  Computer Science Department \\
  Stanford University \\
  \texttt{jiltang@cs.stanford.edu} 
 }

\begin{document}

\maketitle

\begin{abstract}
Generative models of graphs are well-known, but many existing models are limited in scalability and expressivity.  We present a novel sequential graphical variational autoencoder operating directly on graphical representations of data.  In our model, the encoding and decoding of a graph as is framed as a sequential deconstruction and construction process, respectively, enabling the the learning of a latent space. Experiments on a cycle dataset show promise, but highlight the need for a relaxation of the distribution over node permutations.

\end{abstract}

\section{Introduction}	
Graphical representations of data and complex relationships are ubiquitous in myriad fields, from social networks to molecular models, motivating substantial modern interest in machine learning on graphs. Several families of graphical machine learning models exist, falling under broad categories of node-level classification, graph-level classification, and generative models of graphs\cite{wu2019comprehensive}. The latter category is mathematically challenging due to the rich structures representable by graphs but has promising applications in such areas as link prediction\cite{simonovsky2018graphvae} and chemical molecule generation\cite{de2018molgan}.

A large family of graphical generative models, ranging from GraphVAE\cite{simonovsky2018graphvae} to GraphNVP\cite{madhawa2019graphnvp}, generate from a learned distribution of matrix or tensor representations of graphs. While this approach takes advantage of a rich understanding of generative models for vectorized data, it forces the learning over a less natural representation of graphical data and is oftentimes limited to a predetermined node count. GraphRNN\cite{you2018graphrnn} circumvents the latter limitation by emitting a sequence of graph-building steps, but also learns over vectorized representations.

A promising quasi-autoregressive sequential generative model of graphs whose inference structure is based on Graphical CNNs\cite{battaglia2018relational}, and therefore operates on graphical rather than vectorized representations, was presented by Li et al. in \cite{li2018learning}. (For convenience, we will refer to this model as SGGN [sequential graphical generative network].) They demonstrate impressive performance on a chemical molecule generation benchmark. However, their model is not a latent variable model and offers no way of exploring or elucidating the substructure of the learned dataset.

Here we present a novel variational autoencoder based on SGGN which enables the learning of a latent space over a graph dataset. Unlike existing latent variable models of graphs, our generative process takes full advantage of the relational inductive biases expressed by the structure of graphs.

\section{SGNN}

Li et al.\cite{li2018learning} propose SGNN, a generative model of (undirected) graphs based on the graph net architecture\cite{battaglia2018relational}. In this architecture, which has seen use in myriad supervised and semi-supervised tasks\cite{wu2019comprehensive}, computations are performed by \textit{propagating} messages between nodes, updating node and edge vector representations, and updating a global graph embedding after each step. This architecture has shown promising results due to its use of relational inductive biases.

In SGNNs, the results of graph propagation and the global graph embedding are used to probabilistically guide a \textit{sequence of actions} which sequentially construct a graph. The generation process begins with a lone node and its embedding vector, and  adds nodes and edges until the model, based on the graph embedding at that point in time, decides to stop. More formally, the model performs the following computations\cite{li2018learning}:

\begin{algorithm}[H]
\SetAlgoLined
Initialize a single node embedding $\mathbf{h}_0$\\
 \While{TRUE}{
  Perform graph prop $\mathbf{h}_V = \text{prop}(\mathbf{h}_V,G)$\\
  Compute global graph embedding $\mathbf{h}_G = R(\mathbf{h}_V,G)$ \\
  Sample from distribution over new node types  $f_\text{addnode}(G) \sim \text{softmax}(f_{an}(\mathbf{h}_G))$ \\
  
  \If{$f_\textit{addnode}(G) = $ NO NODE}{
   break}
  Initialize new node $v$ with embedding $\mathbf{h}_v = R_\textit{init}(\mathbf{h}_G)$ \\
      \While{sample TRUE from $\text{Bern}(\sigma(f_{ae}(\mathbf{h}_G, \mathbf{h}_v))$}{
      Define a distribution over all $u \in V - \{v\}$ with logits $s_u = f_s(\mathbf{h}_u, \mathbf{h}_v)$\\
      Sample a node $v' \sim \text{softmax}(\mathbf{s})$ \\
      Initialize an edge $\mathbf{h}_{\{v, v'\}} = R_{init\_e}(\mathbf{h}_v, \mathbf{h}_{v'})$ \\
      Perform graph prop $\mathbf{h}_V = \text{prop}(\mathbf{h}_V,G)$\\
      Compute global graph embedding $\mathbf{h}_G = R(\mathbf{h}_V,G)$ \\
    }
 }
 \caption{SGGN}
\end{algorithm}

Here, $\text{prop}$ refers to a typical message-passing operation, in which the each node embedding is updated based on messages from all incoming nodes and the messages are computed from both node vectors and the connecting edge vector; $\mathbf{h}_G$ is an embedding of the overall graph; and $R$ is a reduce operation over all the node embeddings. All functions are parameterized as neural networks with learnable parameters; oftentimes (and in this work), the update operation in prop is a GRU. Also note that multiple (distinct) rounds of prop are often performed between operations.

SGGN, due to its quasi-autoregressive nature, is able to produce log-likelihoods of sampled graphs. However, it is not a latent variable model, as the distribution is defined only by the probabilities of discrete decisions at each time step\footnote{While the authors do admit a conditional SGGN, they do not learn a latent representation.}. Additionally, during training, the ordering of node construction is either fixed to a canonical ordering (which does not permit either rewarding or penalizing generating the graph via other orderings) or it is randomized. This strategy is, however, suboptimal; orderings can convey important information about the graphical structure\cite{vinyals2015order}.

Building upon SGGNs, we present two chief novelties: we learn a latent space while preserving all relational inductive biases and generative procedure of SGNN, and we learn rather than impose or randomize a construction order.

\section{Model}

We propose recasting the sequential graph net generative as the decoder of a VAE, and the seed vector of the initial node $\mathbf{h}_0$ as the latent representation $\bfz$ of the graph. Therefore we model the log-likelihood of a graph $\bfx$ as 
$$p_\theta(\bfx) = \int_\bfz p_\theta(\bfx|\bfz)p_\theta(\bfz) d\bfz$$
where $p_\theta(\bfx|\bfz)$ is given by the SGGN and $p_\theta(\bfz)$ is a prior (in our work, unit Gaussian). We therefore view the SGGN as a complex mapping from a distribution over real-valued vectors $\bfz$ to full-blown graphical representations $\bfx$. Because this SGGN is a decoder that constructs the graph from its initial node representation, we call it the \textit{graph constructor}.

There is an additional nuance imposed by the set of permutations by which $\bfx$ can be generated from $\bfz$, which we shall denote $\pi$. For each $\pi$, there is a distinct and nonzero probability that a structure $\bfx$ can be generated from $\bfz$, since each step in the constructor is probabilistic. Therefore, we actually model
$$p_\theta(\bfx) =  \sum_{\pi} \bigg [ \int_\bfz p_\theta(\bfx, \pi| \bfz)p_\theta(\bfz) d\bfz \bigg ]$$

However, this is highly intractable due to the large space of $\pi$ and the continuous nature of $\bfz$. Therefore, to optimize the log-likelihood we introduce an amortized variational model $q_\phi(\bfz, \pi | \bfx)$, which we use to conduct importance sampling:

$$p_\theta(\bfx) = \mathbb{E}_{\bfz, \pi \sim q_\phi(\bfz, \pi | \bfx)}\left[\frac{p_\theta(\bfx, \pi | \bfz)}{q_\phi(\bfz, \pi | \bfx)}p_\theta(\bfz)\right]$$

We propose modelling $q_\phi$ as a \textit{reverse} SGGN in which a graph $\bfx$ is \textit{deconstructed} according to learned propagation and decision making rules. This operation functions as the encoder of a VAE and we call it the \textit{graph destructor}. Note the under this conception, the $\bfz$ and $\pi$ returned by the destructor are deterministically related; that is, for any $\pi$ and $\bfx$, the posterior $q_\phi(\bfz | \pi, \bfx)$ is a point distribution.

More formally, our graph destructor computes the following:

\begin{algorithm}[H]
\SetAlgoLined
Intake a graph $G = \bfx$ \\
If necessary, initialize node and edge representations $\mathbf{h}_V$, $\mathbf{h}_E$ from node and edge types\footnote{In our implementation, we use a MLP to map the type to the desired embedding dimensionality.}\\
 \While{$\bfx$ has more than one node}{
  Perform graph prop $\mathbf{h}_V = \text{prop}(\mathbf{h}_V,G)$\\
  Define a distribution over all $u \in V$ with logits $s_u = f_r(\mathbf{h}_u)$ \\
  Sample a node $v \sim \text{softmax}(\mathbf{s})$ \\
  Remove node $v$ from the graph
 }
 \caption{Graph destructor}
\end{algorithm}

We return the embedding of the one remaining node as $\bfz$, and the order in which nodes are removed from $G$ corresponds to the inverse of the construction order $\pi$. Note that the destructor does not have a notion of the order in which the \textit{edges} are removed. Therefore, the sequential generative model of SGGN as currently taken will not suffice, as during training it demands a edge generation order as well. We therefore modify the procedure to arrive at the following:

\begin{algorithm}[H]
\SetAlgoLined
Initialize a single node embedding $\mathbf{h}_0 = \bfz$\\
 \While{TRUE}{
  Perform graph prop $\mathbf{h}_V = \text{prop}(\mathbf{h}_V,G)$\\
  Compute global graph embedding $\mathbf{h}_G = R(\mathbf{h}_V,G)$ \\
  \If{sample FALSE from $\text{Bern}(f_{an}(\mathbf{h}_G))$}{
   break}
  Initialize new node $v$ with embedding $\mathbf{h}_v = R_\textit{init}(\mathbf{h}_G)$ \\
  \ForEach {$u \in V - \{v\}$}{
  Sample edge type $f_\text{addedge} \sim \text{softmax}(f_{ae}(\mathbf{h}_u, \mathbf{h}_v, \mathbf{h}_G))$ \\
  \If{$f_\text{addedge}$ is not NO EDGE}{
  Initialize edge $\{u, v\}$ with embedding of type $f_\text{addedge}$
  }
  }
 }
 \caption{Graph constructor}
\end{algorithm}

Here, we have simply replaced the sequential addition of edges, governed by a softmax at each step, with a one-step addition in which all possible edges are independently considered. While this independence assumption limits the expressivity of our model, it makes learning more tractable.

We train $p_\theta$ and $q_\phi$ using teacher forcing and ascend the gradient of the variational lower bound to the log-probability:
$$ \log p_\theta(\bfx) \geq \mathbb{E}_{\bfz, \pi \sim q_\phi(\bfz, \pi | \bfx)}\left[\log \frac{p_\theta(\bfx, \pi|\bfz)}{q_\phi(\bfz, \pi | \bfx)}p_\theta(\bfz)\right]$$

To optimize over $q_\phi$, we find the gradient using the REINFORCE trick:

\begin{align*}
    &\nabla_\phi \mathbb{E}_{\bfz, \pi \sim q_\phi(\bfz, \pi | \bfx)}\left[\log \frac{p_\theta(\bfx, \pi|\bfz)}{q_\phi(\bfz, \pi | \bfx)}p_\theta(\bfz)\right] \\
    &= \nabla_\phi  \sum_{\pi} \Bigg [ \int \log \bigg (\frac{p_\theta(\bfx, \pi|\bfz)}{q_\phi(\bfz, \pi | \bfx)}p_\theta(\bfz) \bigg ) q_\phi(\bfz, \pi | \bfx) d\bfz \Bigg ] \\
    &= \sum_{\pi} \Bigg [ \int \big ( \nabla_\phi [\log (p_\theta(\bfx, \pi|\bfz) p_\theta(\bfz)) q_\phi(\bfz, \pi | \bfx)] - \nabla_\phi [\log (q_\phi(\bfz, \pi | \bfx)) q_\phi(\bfz, \pi | \bfx)] \big ) d\bfz \Bigg ] \\
    &= \sum_{\pi} \Bigg [ \int \big ( \log (p_\theta(\bfx, \pi|\bfz) p_\theta(\bfz)) \nabla_\phi [q_\phi(\bfz, \pi | \bfx)] - \frac{\nabla_\phi [q_\phi(\bfz, \pi | \bfx)]}{q_\phi(\bfz, \pi | \bfx)} q_\phi(\bfz, \pi | \bfx) \\
    &\qquad\qquad\quad - \log (q_\phi(\bfz, \pi | \bfx)) \nabla_\phi [q_\phi(\bfz, \pi | \bfx)] \big ) d\bfz \Bigg ] \\
    &= \sum_{\pi} \Bigg [ \int \nabla_\phi \big [q_\phi(\bfz, \pi | \bfx) \big ] \big [ \log(p_\theta(\bfx, \pi|\bfz) p_\theta(\bfz)) - 1 - \log (q_\phi(\bfz, \pi | \bfx)) \big ] d\bfz \Bigg ] \\
    &= \sum_{\pi} \Bigg [ \int \nabla_\phi \big [q_\phi(\bfz, \pi | \bfx) \big ] \bigg [ \log \bigg (\frac{p_\theta(\bfx, \pi|\bfz)}{q_\phi(\bfz, \pi | \bfx)}p_\theta(\bfz) \bigg ) - 1 \bigg ] d\bfz \Bigg ] \\
    &= \sum_{\pi} \Bigg [ \int q_\phi(\bfz, \pi | \bfx)  \nabla_\phi \big [\log q_\phi(\bfz, \pi | \bfx) \big ] \bigg [ \log \bigg (\frac{p_\theta(\bfx, \pi|\bfz)}{q_\phi(\bfz, \pi | \bfx)}p_\theta(\bfz) \bigg ) - 1 \bigg ] d\bfz \Bigg ] \\
    &= \mathbb{E}_{\bfz, \pi \sim q_\phi(\bfz, \pi | \bfx)} \Bigg [ \nabla_\phi \big [\log q_\phi(\bfz, \pi | \bfx) \big ] \bigg [ \log \bigg (\frac{p_\theta(\bfx, \pi|\bfz)}{q_\phi(\bfz, \pi | \bfx)}p_\theta(\bfz) \bigg ) - 1 \bigg ] \Bigg ]
\end{align*}
Therefore we can estimate the gradient $\nabla_\phi \text{ELBO}$ via Monte Carlo sampling with  
\begin{align*}
    \frac{1}{n} \sum_{i=1}^n \Bigg [ \nabla_\phi \big [\log q_\phi(\bfz^{(i)}, \pi^{(i)} | \bfx) \big ] \bigg [ \log \bigg (\frac{p_\theta(\bfx, \pi^{(i)}|\bfz^{(i)})}{q_\phi(\bfz^{(i)}, \pi^{(i)} | \bfx)}p_\theta(\bfz^{(i)}) \bigg ) - 1 \bigg ] \Bigg ]
\end{align*}
where $\bfz^{(i)}, \pi^{(i)} \sim q_\phi(\bfz, \pi | \bfx)$. While it is generally infeasible to evaluate $q_\phi(\bfz, \pi | \bfx)$ and $p_\theta(\bfz, \pi | \bfz)$, we can directly do so for the elements that we sample by calculating the probability of each step of the deconstruction and construction.

\section{Results}



We perform experiments on a toy cycle dataset as a proof-of-concept of our architecture. Specifically, we construct a 100-graph dataset composed of cycles of length 5-14. Since we define only one edge type and only one node type, the edge and node initializations in the destructors are constant. We use a 5-dimensional node embedding for graph prop and therefore learn a 5-dimensional latent space over $\bfz$.

We train using mini-batch gradient descent on the mean negative ELBO using an Adam optimizer and a learning rate of $\alpha=0.01$. We evaluate the generation accuracy of the model by sampling 100 samples from the latent distribution $p_\theta(\bfz)$, passing them through the constructor, and determining the fraction of valid cycles. The negative ELBO and generation accuracy over a sample training cycle are shown in Figure 1.

\begin{figure}[H]
    \centering
    \includegraphics[width=\textwidth]{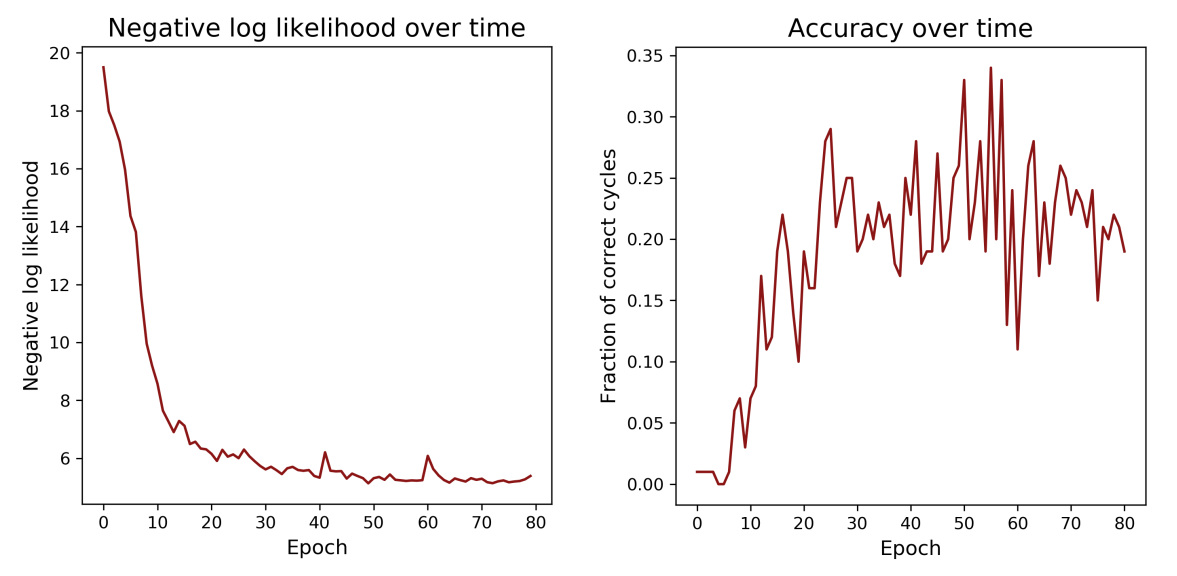}
    \caption{Mean minibatch negative log likelihood (approximated as negative ELBO) and valid cycle generation accuracy over time. Note that while the loss appears to continue decreasing, the generation accuracy peaks at around epochs 50-60.}
\end{figure}

We achieve lower perplexity (below 5.1) on other training runs, but not higher generation accuracy. We posit this discrepancy may be attributed to the looseness of our lower-bound approximation. We also observe that training is highly sensitive to parameter initialization: due to the large and discrete space of permutations $\pi$, a substantial proportion of training runs failed to learn a canonical $\pi$ for any $\bfx$ because the decoder would rarely revisit the same $\pi$, thus making $q_\phi(\bfz|\bfx)$ multimodal and $p_\theta(\bfx | \pi, \bfz)$ difficult to optimize. Altogether, compared to SGNN, we achieve lower perplexity on the cycle dataset but lower generation accuracy\cite{li2018learning}.

\begin{figure}[H]
    \centering
    \includegraphics[width=0.95\textwidth]{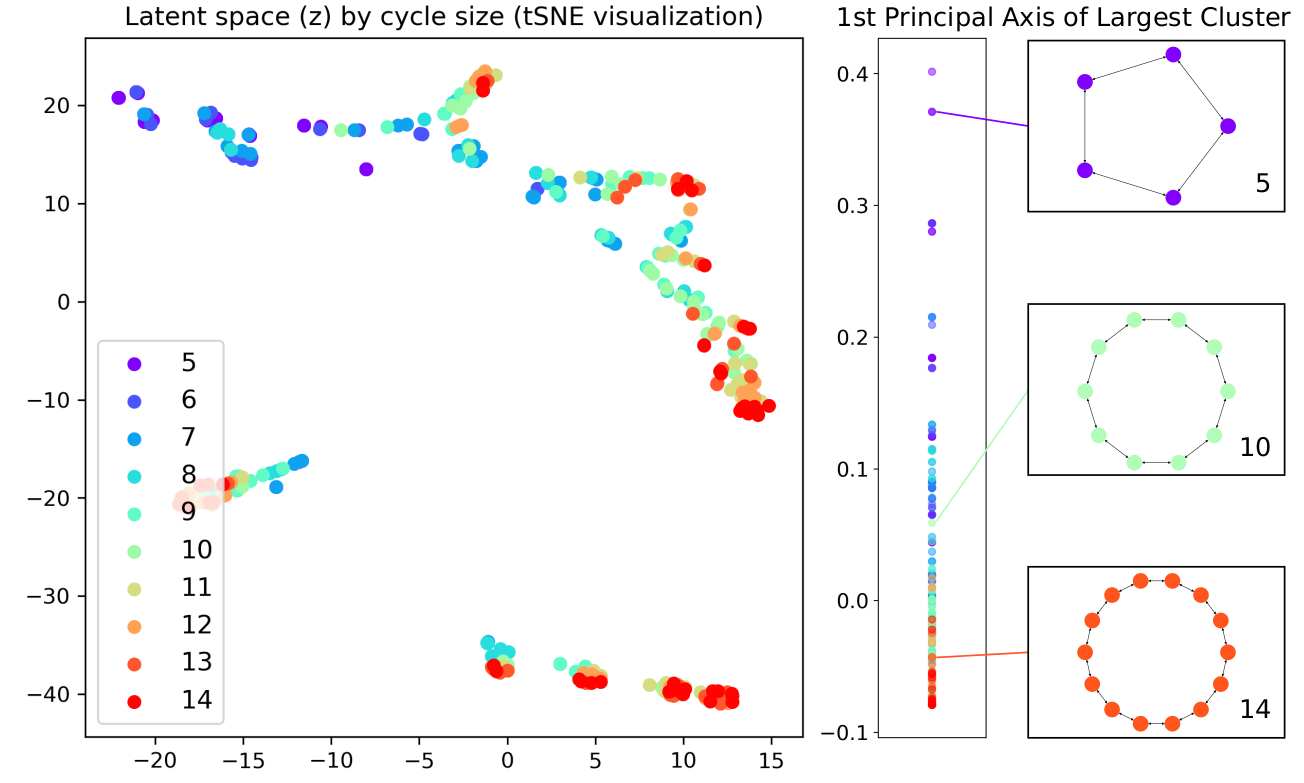}
    \caption{tSNE visualization of 400 points in the 5-dimensional latent space obtained by sampling $q_\phi(\bfz, \pi | \bfx)$. Note the multiple clusters representing multiple destruction modalities, the linearity of the cluster, and the interpretability of the principal axis obtained by PCA.}
\end{figure}

We evaluate the learned latent space (the primary novelty of our method) by first visualizing the sampled latent space embeddings of 400 graphs (40 for each cycle length in 5 to 14) with tSNE in Figure 2. The destructor network has learned a semantically meaningful latent space, with a principal axis of variation corresponding to cycle length. Interestingly, however, it has not learned to encode graphs into a single region of the latent space; instead, there are multiple clusters, each of which displays a interpretable principal axis. (In Figure 2, 70\% of the data points are in the largest cluster.)

We posit that the performance of the generative model is contingent on locating most embeddings in a single cluster, as it is difficult to fit a multimodal posterior $q_\phi(\bfz | \bfx)$ to the Gaussian prior $p_\theta(\bfz)$ in minimizing the KL-term of the ELBO $$D_\text{KL}(q_\phi(\bfz|\bfx)||p_\theta(\bfx)).$$
In addition to complicating loss minimization, a multi-clustered latent representation ensures the latent space is sparse, which further explains the low generative accuracy of even the best training runs. Unfortunately, there is not an obvious way to force convergence to a single cluster during training. Nevertheless, we note that the destructor has learned to cluster representations in a relatively small number of destruction modalities; because $\pi$ and $\bfz$ are deterministically related, each point $\bfz$ corresponds to a unique way of deconstructing the graph $\bfx$. We see that this cluster of modalities is opposed by combinatorial growth of the space of $\pi$, which manifests as a branching effect in the latent space clusters\footnote{As shown in the PCA, this branching does not end up complicating the principal axis of variation.}.

A further desirable quality of a latent space is that decoding interpolated representations should lead to a semantically meaningful progression in the generated space. We perform this evaluation by taking latent representations regularly spaced along the first principal axis of the largest cluster and evaluating the distribution of graphs $\bfx$ generated under $p_\theta(\bfx|\bfz)$. These results are shown in Figure 3.

\begin{figure}[H]
    \centering
    \includegraphics[width=0.7\textwidth]{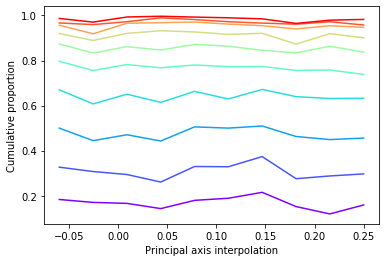}

    \caption{The distribution of cycle lengths is plotted as a cumulative proportion (i.e., fraction of cycles with length no greater than the number associated with each line) for 10 regularly spaced points interpolating the first principal axis of the largest cluster in the latent space. Proportions are computed from 1000 graphs sampled from $p_\theta(\bfx|\bfz)$. Colors correspond to the key in Figure 2.}
\end{figure}

As the data shows, we unfortunately do not find a compelling trend between the distribution of cycle lengths when interpolating along the most compelling axis. A quantitative demonstration of this fact is given in Table 1. We posit that the encoder network has learned to ignore the latent representation $\bfz$ in generating graphs, a common occurrence for expressive models\cite{zhao2017infovae}. Adding a mutual information term, as in InfoVAEs\cite{zhao2017infovae}, may help address this phenomenon. However, a more serious underlying cause may be associated with the multimodal amortized posterior $q_\phi(\bfz|\bfx)$, at least in the early stages of training, as discussed above, which would have hindered the graph constructor from learning a meaningful correspondence from $\bfz$ to $\bfx$.

\begin{table}[H]
    \centering
    \caption{Mean cycle length ($N<=1000$) at interpolated locations along the principal axis.}
    \begin{tabular}{ccccccccccc} \hline
    \textbf{Coordinate} & -0.06 &-0.025 & 0.008 & 0.043 & 0.077 & 0.112 & 0.146 & 0.181 & 0.215 &
0.25 \\ \hline
    \textbf{Mean} &  7.7 & 7.98 & 7.81 & 7.92 & 7.695 & 7.797 & 7.652 & 8.056 & 7.989 & 8.003 \\
    \textbf{Std. err.} & 0.008 & 0.010 & 0.008 & 0.008 & 0.008 & 0.009 & 0.010 & 0.009 & 0.008 & 0.009 \\
    \hline \end{tabular}
\end{table}

\section{Discussion and conclusion}

We have presented SGVAE, an graph autoencoder based on sequential operations on graphical representations. While the architecture and projected latent embeddings of SGVAE are promising, we recognize that our generative model was unable to learn a latent space in which interpolation is meaningful, at least in these experiments.

We note that the training of the model, even on a toy dataset, was difficult due to the high variance of the gradient $\nabla_\phi$, compounded by the ill-formalized learning task over $\pi$. To address the former, we propose utilizing control variates in future experiments. The latter issue is significantly trickier. Although methods for relaxing distributions on permutation spaces exist (i.e. the Plackett-Luce distribution), such methods are ill-suited because $q_\phi(\pi | \bfx)$ is not explicitly parameterizable; indeed, it cannot even be disentangled from $q_\phi(\bfz|\bfx)$.

We instead posit the following relaxation over graphical representations as a component of the graph destructor: instead of permanently removing nodes, we can down-weight the nodes based on their removal probability, but still include them in subsequent graph prop operations. The encoding is still taken to be the embedding of the final node to be formally removed, but now the gradient is able to propagate to the removal probabilities rather than stop at discrete decision-making steps. Additionally, whereas currently the discrete graph operations tend to divide the distribution of $q_\phi(\bfz|\bfx)$, under the proposed relaxation the distribution should be much more well-behaved.

Finally, the independent edges assumption also deserves reconsideration, especially for more complex datasets, but it is unlikely to be the cause of the performance seen in these experiments.

\section{Code}
Our implementation of SGVAE and results and analysis of experiments may be found at \url{https://github.com/bjing2016/sgvae}.




\bibliographystyle{unsrt}

\bibliography{references}

\end{document}